\documentclass[11pt]{article}
\usepackage[a4paper,margin=1in]{geometry}
\usepackage{fontspec}
\usepackage{microtype}
\usepackage{booktabs}
\usepackage{array}
\usepackage{tabularx}
\usepackage{caption}
\usepackage{amssymb}
\usepackage{enumitem}
\usepackage[hidelinks]{hyperref}
\title{\vspace{-3em}\bfseries\Large Artificial Epanorthosis\\[0.35em]
  \normalfont\large Why large language models overuse a classical rhetorical figure, and how to mitigate it}
\author{Federico Boggia}
\date{July 2026}
\begin{document}
\maketitle

\begin{abstract}
A rhetorical figure that Cicero and Quintilian catalogued two thousand years ago reappears, systematically, in the text of large language models: \emph{epanorthosis}, the self-correction of the specimen «This is not a course. It is a journey of transformation». This essay argues that the overuse is a trained disposition, driven mainly by a training distribution rich in promotional prose and by preference tuning (RLHF) that rewards confident, emphatic phrasing; the left-to-right nature of generation is an amplifier rather than the root cause. Building on evidence that models diverge from human rhetorical style, and on Fontanier's classification of epanorthosis as a figure of \emph{thought}, it sets out a programme that scores the figure against genre-specific human baselines through an \emph{Epanorthosis Index} (density relative to the human rate). A first measurement, on three sizes of one instruction-tuned model family, finds mis-calibration by register in both directions: the models overshoot in oratory (about twofold, near threefold in Italian, concentrated in the larger tiers) and undershoot in informal question-and-answer writing, while matching humans in argument, journalism, and encyclopedic prose. Three constructive contributions follow: a survey of mitigation techniques centred on lightweight LoRA adapters; a demonstration, in Italian, that a one-line instruction cuts the figure by half to nearly three-quarters and that a supervised-fine-tuning adapter removes it almost entirely, with a scaling coefficient that dials the reduction back onto the human rate; and the argument that the target is \emph{calibration to the human rate for each genre}, not elimination. It closes on the stakes: the real risk is that we begin to write like the machines.
\end{abstract}

\medskip\noindent \emph{Keywords:} epanorthosis, rhetoric, large language models, autoregressive generation, RLHF, LoRA, controllable text generation, AI-text detection

\section*{1. Introduction}

Read the next answer a chatbot gives you, or the next motivational post that scrolls through a professional feed, and a structure begins to surface that, once seen, is difficult to unsee:

\begin{quote}
«I am not a consultant. I am an architect of change.»\\
    «We do not sell products. We build experiences.»\\
    «This is not a course. It is a journey of transformation.»
\end{quote}
The model asserts something and, in the same breath, negates it in order to replace it with something larger, stronger, more solemn. The tic is so pervasive that it has become a kind of signature of automatically generated writing. What is rarely said is that the tic has a name: a Greek name, more than two thousand years old, for a device Cicero and Quintilian wielded with full awareness. That name is \emph{epanorthosis}.

The observation has empirical support. A recent large study comparing parallel corpora of human- and model-written text found that language models systematically diverge from human writers precisely at the level of grammatical and rhetorical style, that the divergence is \emph{larger} for instruction-tuned models than for base models, and that it does not vanish as models scale up.\cite{r1} The rhetorical fingerprint, in other words, is being reinforced by the very training stages meant to make models more useful, and it persists as models grow.

This essay pursues a question that is simple to pose and surprisingly hard to close. When a machine that knows nothing of rhetoric spontaneously reproduces a classical rhetorical figure, what is actually happening? Is it a coincidence, a meaningless statistical artefact? Or does the convergence between ancient rhetoric and the statistics of neural networks tell us something about language itself? The argument proceeds in three movements. Sections 2--3 characterise the figure and trace where the overuse comes from. Sections 4--6 turn the phenomenon into an object of measurement and ask whether it is a figure of rhetoric or an artefact of prediction. Section 7, the core contribution, surveys concrete future directions for mitigating the overuse, organised around LoRA adapters and other lightweight techniques. A note on method: the paper tries to practise what it argues, keeping its own prose free of gratuitous epanorthosis, and the corrective turns it does use are the legitimate, content-bearing kind it defends in Section 7.

\section*{2. Epanorthosis in classical rhetoric}

The word derives from the Greek epan\'orth\=osis (\emph{epanorthosis}): \emph{epi}, ``upon, in addition'', and \emph{anorthosis}, ``a straightening'', meaning literally ``a further straightening''. The speaker returns to what has just been said in order to correct it. The correction works as a deliberate rhetorical move. One goes back to reinforce, to soften, or to make more precise. In Latin the figure travels under the name \emph{correctio}. Three functional variants are worth separating, along the lines of the tripartition Fontanier would later draw (strengthening, softening, retracting). There is the \emph{emphatic} correction, an upgrade in which a term is replaced by a stronger one (``he is good, indeed excellent''); the \emph{attenuative} correction, a downgrade (``it is a disaster, well, a setback''); and the properly \emph{corrective} variant, which changes the substance (``he was running, or rather, walking briskly''). It is one of the most common figures of everyday speech: every ``I mean'', ``rather'', ``or better'' is an act of epanorthosis performed without noticing.

The rhetorical tradition recognised the force of self-correction early. In Book IX of the \emph{Institutio Oratoria} (c. 95 CE), Quintilian discusses the device under several related Latin terms.\cite{r4} In a catalogue taken from Cicero, \emph{correctio} may occur before or after an utterance, or when the speaker rejects something previously said (\emph{Institutio Oratoria} 9.1.30). Elsewhere Quintilian gives examples of \emph{emendatio} and \emph{reprehensio}, in which an expression is revised, qualified, or replaced (9.2.17--18), and distinguishes corrections affecting the thought from those affecting the wording (9.3.88--89). Particularly relevant is his discussion of \emph{paenitentia dicti}, the simulated regret or reconsideration of something just said: such devices make discourse appear simple and unprepared and therefore make the speaker seem less calculated to the judge (9.2.59--60). Quintilian thus presents self-correction as a flexible rhetorical operation capable of revising an expression and staging spontaneous thought, although he does not formulate a fixed three-part taxonomy of its functions. A clear Ciceronian instance is in the First Catilinarian, where a flat assertion is corrected upward in the same breath:\cite{r5}

\begin{quote}
Hic tamen vivit. Vivit? \emph{immo vero etiam in senatum venit.}\\[3pt]{\small Cicero, \emph{In Catilinam} I.2: ``Yet this man lives. Lives? Nay, he even comes into the Senate.''}
\end{quote}
The particle \emph{immo} (``nay, rather'') is the Latin correction-marker par excellence: the first claim is retracted as an understatement and replaced by a stronger one. Epanorthosis proper should be kept apart from mere accumulation. Cicero's more famous ``Abiit, excessit, evasit, erupit'' (\emph{In Catilinam} II.1), four verbs that intensify without retracting one another, is a different figure altogether: Quintilian cites the line among the accumulations Caecilius took for pleonasm (9.3.45--46) and again under \emph{homoeoteleuton} (9.3.77), and his \emph{gradatio} has each step take up the preceding word (9.3.54--55), which these four verbs never do. The most devastating vernacular case of genuine corrective epanorthosis may be Dante's, at the close of the Ugolino canto: ``Poscia, più che 'l dolor, poté 'l digiuno''.\cite{r6} There the final correction reverses the meaning of the entire scene, revealing that hunger, more than grief, killed Ugolino.

For the present argument the decisive author is a nineteenth-century French rhetorician, Pierre Fontanier, in \emph{Les Figures du discours} (1827).\cite{r2} Fontanier draws a distinction that seems pedantic yet turns out to be crucial: he separates \emph{correction}, which he treats as a figure of style, from \emph{épanorthose} (or \emph{rétroaction}), which he places among the \textbf{figures of thought}. In his own words, here in English translation: ``Rétroaction, otherwise Epanorthosis, which must not be confused with Correction, a figure of style, consists in returning to what one has said, either to strengthen it, or to soften it, or even to retract it entirely.'' The consequence is far-reaching. If epanorthosis is a figure of \emph{thought} rather than of style, then it is a general discursive operation. It can be observed wherever there is language, including the output of a statistical model, and it extends beyond fine writing or oratory; crucially, it can be studied independently of the intention of whoever produces it. This is the bridge that lets us pass from Cicero to a chatbot without cheating. Standard reference works confirm the classification: Lausberg's \emph{Handbook of Literary Rhetoric} catalogues \emph{correctio} among the figures that stage a revision of the utterance.\cite{r7}

\section*{3. Where the overuse comes from}

The overuse is best understood as a trained disposition, and its main sources lie in how models are built rather than in how they emit tokens. The first is the \textbf{training distribution}: vast quantities of web text, a substantial fraction of which is copywriting, motivational posts, and landing pages, all genres in which emphatic epanorthosis correlates with engagement. This is a plausible pressure rather than a measured one, and quantifying it corpus-side is itself part of the research programme of Section 4. The model learns that the structure works. The second source, and probably the decisive one, is \textbf{reinforcement learning from human feedback} (RLHF): the alignment stage in which annotators rank candidate responses.\cite{r10} Raters tend to prefer answers that sound confident, incisive, and committed, and few things sound as confident as a well-placed upward correction. There is empirical support for this link: investigating why words such as ``delve'' and ``intricate'' became overrepresented in model output, Juzek and Ward found no evidence that architecture, decoding algorithm, or the raw training data were responsible, and their comparative model testing is consistent with \emph{RLHF} playing a role, though they stop short of identifying it as the source and leave the question open.\cite{r11} If preference tuning can inflate individual words, it can inflate a whole construction. This attribution is inferential: it extrapolates from a lexical study to a construction, and the measurements of Section 4 (human against instruction-tuned output, with no base-model-versus-aligned comparison) cannot themselves adjudicate the cause.

Generation itself adds a weaker, secondary pressure, and it is worth being precise about how much. A transformer language model produces text \textbf{autoregressively}, one token at a time, left to right, and a committed token cannot be retracted.\cite{r8} It is tempting to conclude that the model therefore repairs by correcting forward, ``good, indeed excellent'', which is the surface shape of epanorthosis. That inference is too strong: a well-trained model usually emits the intended token on the first attempt, so forced forward-correction is the exception, not a structural necessity. What the left-to-right constraint does contribute is subtler. Unlike a human writer, who drafts, deletes, and reveals only a clean result, the model cannot revise before publishing, so whatever emphatic-correction disposition it has acquired from training surfaces in the open, unedited, instead of being smoothed away. Decoding plays a similar amplifying role: likelihood-maximising and top-\emph{k} decoding yield text that is fluent yet ``too probable'', short on the diversity of human writing,\cite{r9} and the ``Not X. Y'' frame is one such high-probability path.

The surface forms are a small, recognisable family. The commonest is ``Not X. Y'': negate a term and replace it with a grander one. (The specimens that open this essay are the fastest proof that the pattern can be recognised.) The variants are familiar once named: ``Not only X, but Y''; ``X, or rather, Y''; ``It is not X, it is Y''; ``X? No. Y''. A useful field heuristic: three or more such frames in a short passage is a strong indicator of generated text, though never a conclusive one.

\section*{4. Measuring the phenomenon}

A signature one can only feel is of limited scientific use. Can the overuse be measured systematically, and compared against a human baseline? Recent work suggests it can, and that the effect is real. The excess-vocabulary method of Kobak and colleagues, applied to more than fifteen million biomedical abstracts, quantifies post-2022 shifts in word frequency attributable to model-assisted writing and estimates that at least 13.5 percent of 2024 abstracts were processed with a language model, with the share reaching some 40 percent in individual subcorpora.\cite{r12} Liang and colleagues, using a maximum-likelihood estimator over expert- and model-written reference texts, found that between roughly six and seventeen percent of the text submitted as peer reviews at major AI conferences after the release of ChatGPT could have been substantially model-modified.\cite{r13} And, as already noted, Reinhart and colleagues moved the analysis from single words to grammatical and rhetorical features (Biber's feature set) and showed that the stylistic gap is structural, larger for instruction-tuned models, and persistent across scale.\cite{r1} These results establish that rhetorical divergence between human and model text is measurable; epanorthosis intensity is one natural coordinate within that space.

Two families of detector are relevant. The first exploits token-level statistics: GLTR visualises how ``expected'' each token is under a reference model,\cite{r14} and DetectGPT tests whether a passage sits at a local maximum of the model's log-probability, a curvature signature of machine text.\cite{r15} Comparison corpora such as HC3 supply paired human and model answers for training and evaluation.\cite{r16} These methods target \emph{authorship} rather than a specific figure; epanorthosis is one feature among many that they implicitly exploit. The second family targets rhetorical \emph{constructions} directly. Dubremetz and Nivre showed that repetition-based figures (chiasmus, epanaphora, epiphora) can be detected with supervised models that separate rhetorically effective repetitions from accidental ones,\cite{r18} and Bothwell and colleagues formalised \emph{rhetorical parallelism detection} as a task with datasets, metrics, and baselines.\cite{r19} Epanorthosis works by substitution where those figures work by repetition, yet it too is a construction rather than a bag of words, so its detection belongs to this second, construction-level tradition.

A methodological caveat frames what follows. Fontanier's licence to study epanorthosis regardless of intention comes from treating it as a figure of \emph{thought}; yet any lexical or construction-level detector measures its \emph{stylistic} manifestation, the surface ``Not X. Y'', which is the figure of style Fontanier keeps separate. The protocol below therefore measures a proxy: the observable trace of a discursive operation, not the operation itself. On this footing one can sketch a concrete measurement protocol. First, build a \textbf{human baseline corpus} stratified by genre: spontaneous speech (for Italian, the KIParla corpus of transcribed spoken interaction offers a genre where discourse markers are frequent and the figure often serves merely as a filler),\cite{r17} academic prose, nineteenth- and early twentieth-century narrative, journalism, and social writing. Draw it, wherever possible, from material predating the diffusion of generative models, so that the human rate is measured before contamination. Second, run a two-stage detector: a high-recall lexical pass over corrective markers (\emph{but}, \emph{rather}, \emph{actually}, \emph{I mean}, the ``not\ldots{}but'' frame), followed by a construction-level classifier that filters the many false positives; words like ``but'' and ``I mean'' recur constantly without any corrective function, especially in speech, where they serve largely as fillers. A first-pass implementation of the ``not\ldots{}but'' frame alone, the marker-based triggers of the wider pass being still unimplemented, evaluated against a 206-window hand-annotated Italian gold standard (two annotators; Cohen's $\kappa$ = 0.35 on the candidate labels), is a high-recall, low-precision proxy (micro precision 0.45, recall 0.52). Its precision is decisively source-dependent --- 0.82 on model-generated text, where the figure takes its canonical ``not X, but Y'' form, but only 0.17 on human text, where the same markers are overwhelmingly ordinary adversatives rather than corrections --- so the detector is most reliable precisely on model output and noisiest on the human baseline (per-genre and per-source figures in Appendix B). The lesson is methodological: at the tail, the phenomenon demands a human judgement of function that no lexical trigger supplies on its own. The measurement target, moreover, is frequency \emph{conditioned on genre and communicative purpose} rather than raw frequency; this point returns, decisively, in Section 7.

\subsection*{4.1 A preliminary measurement}

To move from protocol to a first result, I ran an English-language study with an English instantiation of the detector above (distinct from the Italian corpus just described). The human baseline uses public-domain and openly licensed human texts, one to several works per genre, with pre-2022 filtering where feasible: encyclopedic prose (Wikipedia), journalism (Wikinews), academic abstracts (arXiv), narrative fiction (Conan Doyle), informal question-and-answer writing (high-voted pre-2022 Stack Exchange answers), argument (Mill, \emph{On Liberty}), and oratory (inaugural and commemorative addresses, among them Lincoln, Jefferson, and T. Roosevelt). The model sample comprises short texts from neutral, genre-specific prompts that never mention correction or any rhetorical figure, generated across three model sizes from a single instruction-tuned family (Claude Haiku, Sonnet, and Opus, which share one alignment pipeline) and eight genres (seven with a human baseline, plus a model-only promotional register). Every text, human and model, is cut into windows of a few hundred words; density counts the emphatic-correction family (``not X, but Y'', ``not only X but Y'', and the sentence-split ``Not X. Y'') per ten thousand words, and significance is a label-permutation test on the per-window densities.

\begin{table}[ht]\centering
\caption*{\textbf{Table 1.} Emphatic-epanorthosis density (per 10{,}000 words), human vs.\ three model sizes from one instruction-tuned family, over few-hundred-word windows ($n$ = windows). $p$ from a 5{,}000-iteration permutation test; bold marks $p \le 0.05$. The EI column is computed from unrounded densities and can differ marginally from the ratio of the rounded density cells shown (e.g.\ journalism).}
\small
\begin{tabular}{@{}lllll@{}}
\toprule
Genre & Human ($n$) & LLM ($n$) & EI (LLM/human) & $p$ \\
\midrule
Encyclopedic & 1.2 (21) & 1.4 (18) & 1.2$\times$ & 1.00 \\
Journalism & 2.4 (128) & 2.0 (28) & 0.9$\times$ & 0.89 \\
Academic abstract & 3.6 (47) & 7.8 (18) & 2.2$\times$ & 0.28 \\
Narrative fiction & 7.5 (84) & 4.1 (18) & 0.5$\times$ & 0.38 \\
Informal Q\&A & 8.2 (75) & 1.3 (21) & 0.2$\times$ & \textbf{0.05} \\
Argumentative essay & 17.2 (71) & 22.0 (18) & 1.3$\times$ & 0.37 \\
Oratory & 14.9 (19) & 33.5 (17) & 2.2$\times$ & \textbf{0.03} \\
Promotional & n/a & 26.4 (18) & model-only & n/a \\
\bottomrule
\end{tabular}
\end{table}
Two effects reach significance, and they point in opposite directions. In the \emph{oratorical} register the models \emph{overshoot}, producing emphatic self-correction at roughly twice the pooled human rate (33.5 against 14.9 per ten thousand words, \emph{p} = 0.03), an effect carried by the larger tiers (Sonnet and Opus at 37 and 46) while the smallest sits at the human rate (Haiku, 15). In informal \emph{question-and-answer} writing they \emph{undershoot}, falling to about a fifth of the human rate (1.3 against 8.2, \emph{p} = 0.05), where human writers hedge and correct themselves freely and the models stay flat. The remaining registers show no significant gap: argument, encyclopedic prose, and journalism sit at parity, while narrative trends toward model undershoot (EI 0.5$\times$) and academic abstracts toward overshoot, neither reaching significance.

The picture is one of \emph{mis-calibration by register in both directions}: the models pour the figure on where it flatters a persuasive audience and withhold it where a peer would qualify a claim. A specificity check supports reading this as targeted rather than as general verbosity: across genres the models use \emph{fewer} neutral connectives than humans (in oratory, 1.5 against 7.5 per ten thousand words), so the oratorical spike is not mere marker-heaviness. Narrative makes the same point from the other side, since against a proper multi-work baseline the human rate is itself high (7.5) and the models fall below it.

These ratios can be read as a single metric. Define the \textbf{Epanorthosis Index} of a text or corpus in genre \emph{g} as its emphatic-epanorthosis density divided by the human baseline for that genre, EI\textsubscript{g} = D\textsubscript{g} / H\textsubscript{g}: an index of 1 is human-like calibration, above 1 is overshoot, below 1 undershoot. The ``LLM / human'' column of Table 1 is exactly this index, and its spread is the point: EI $\approx$ 2.2 in oratory and $\approx$ 0.2 in informal question-and-answer, a two-sided mis-calibration that a single raw-frequency number would hide. A corpus-level \emph{calibration error}, the mean of |log EI\textsubscript{g}| across genres, condenses this into one number for how far a model sits from human calibration in either direction (0 = perfectly calibrated).

Caveats bound these numbers. The corpus is modest and English-only; the human argumentative, oratorical, and narrative samples are nineteenth- and early twentieth-century, so register is matched but era is not (the encyclopedic, journalistic, academic-abstract, and question-and-answer material is contemporaneous); the detector captures the ``not X, but/. Y'' family, a central but partial slice of epanorthosis; and promotional text has no public-domain human baseline. The per-genre \emph{p}-values, moreover, are not corrected for multiple comparisons across the seven simultaneous tests, so under a strict correction neither of the two marginal results (oratory at 0.03 and informal question-and-answer at 0.05) would survive, and they are reported here as suggestive rather than confirmed. The result is a first signal rather than a settled measurement. It points, however, the same way as this paper's argument: the models' failure is one of appropriateness rather than raw frequency, concentrated in the register where the figure most flatters, which is exactly what Section 7.7 proposes to target. The corpus and detector script are available from the author for replication.

\section*{5. When self-correction does real work}

So far epanorthosis has figured as a stylistic tic. There is also a regime in which the model's self-correction stops being decoration and becomes function. Since 2022 we have known that models answer far better when asked to reason step by step (the chain-of-thought technique),\cite{r20} and that a single instruction such as ``let's think step by step'' elicits the behaviour zero-shot.\cite{r21} In the intermediate steps a fully-fledged corrective epanorthosis appears: ``Wait, that's not right, let me recompute''; ``Actually, let me reconsider.'' Iterative refinement methods make the move explicit, feeding a model its own draft and self-critique.\cite{r22}

The obvious question is whether, when a model writes ``wait, let me reconsider'', it is really reconsidering. The honest answer is uncomfortable. There is no internal module that fires on ``wait'', no representation of ``changing one's mind''. Yet the token sequence ``wait, let me reconsider'' shifts the context, and thereby shifts the probabilities of subsequent tokens toward alternative lines of reasoning. Textual self-correction can therefore \emph{work} even without a mind behind it: the rhetorical form, emptied of mental content, still produces concrete effects. The claim must be qualified, however: Huang and colleagues showed that models often \emph{cannot} reliably self-correct reasoning without an external signal, and may even degrade their answers when asked to revise blindly.\cite{r23} Self-correction is a genuine capability only when the correction is grounded in something concrete, such as a tool result, a verifier, or a retrieved fact, rather than in the mere performance of reconsidering.

There is a historical inversion here worth savouring. Classical rhetoric taught orators to \emph{simulate} self-correction in order to seem more credible; modern engineering teaches machines to \emph{practise} self-correction in order to be more accurate. Same form, opposite direction. Anthropic's Constitutional AI, in which a model drafts, critiques its draft against a set of written principles, and rewrites, is exactly this: engineered epanorthosis, performed behind the scenes before we ever see the answer.\cite{r24} The figure that Cicero staged for the Senate has become an optimisation loop.

\section*{6. Figure of rhetoric, or artefact of prediction?}

When a human says ``good, indeed excellent'', a judgement lies behind the correction; when a model produces the same structure, behind it there is only a frequency in the data. This is Searle's Chinese Room applied to rhetoric:\cite{r25} flawless figures with, apparently, no intentional persuasion behind them. The functionalist reply is that if rhetoric is defined by its effects on a listener rather than by intention, a model's epanorthosis persuades and emphasises exactly as Cicero's does. Both positions are defensible, and the honest verdict is that the categories built for intentional human discourse do not yet settle the case. What matters for the rest of this essay is narrower and practical: whatever its ontological status, the figure is produced in excess, that excess is measurable, and where it is gratuitous it can be turned down. If we wished to do so, could we, and how?

\section*{7. Future directions for mitigation}

This section is the paper's constructive core. Before surveying techniques, one framing decision must be made explicit, because it constrains every method that follows. The objective is to \textbf{calibrate} epanorthosis to context. Correction is often a virtue: in argumentative prose it carries clarity and intellectual honesty, and forcing it to zero would flatten the very texts we value. What matters is the figure's \emph{appropriateness to context}: an emphatic upgrade that suits a keynote reads as grating in a discharge summary. The pilot of Section 4 makes this concrete: in English the models overshoot specifically in the persuasive and oratorical register while matching humans in argument. Mitigation should therefore be register-aware and, above all, calibrated to the human rate for each genre --- turning the figure down where it overshoots (steeply in oratory, and, in the Italian experiments of Sections 7.2 and 7.8, in argument too) without driving the legitimate, content-bearing correction below the human rate. The engineering target, therefore, is a controllable reduction of gratuitous, upward-emphatic epanorthosis at fixed semantic content, ideally exposed as a dial. The survey below moves from the most invasive interventions (retraining weights) to the least (prompting), noting for each its cost, its controllability, and its principal risk. Table 3 summarises the comparison.

\subsection*{7.1 Data-side interventions}

The cheapest place to act is the data, before any gradient is computed. Two moves are available. The first is \emph{counterfactual data augmentation}: mine a corpus for emphatic ``Not X. Y'' constructions and pair each with a de-emphasised paraphrase that preserves content (``I am not a consultant. I am an architect of change'' $\rightarrow$ ``I work as a consultant, with a focus on organisational change''). These pairs become supervision for later stages. The second is \emph{instruction-data curation}: because the divergence is amplified by instruction tuning,\cite{r1} pruning promotional and template-heavy examples from the instruction mixture attacks the problem at its documented source. Data-side methods are attractive because they are non-invasive and compose with everything downstream; their limitation is that they act only at the next training run and offer no inference-time control.

\subsection*{7.2 A LoRA adapter for rhetorical style}

The most promising route, and the one this essay emphasises, is a lightweight fine-tune with \textbf{Low-Rank Adaptation (LoRA)}.\cite{r26} LoRA freezes the base model's weights and injects, in parallel with selected weight matrices (typically the attention query/value projections and the MLP projections), a trainable low-rank update $\Delta$W = BA, whose inner dimension \emph{r} is far smaller than the model dimension. Only B and A are trained; the number of trainable parameters drops by three to four orders of magnitude, and, decisively for this application, the adapter is a small, separable artefact that can be attached or detached at will. One trains a \emph{behaviour} rather than a new model.

Concretely, an ``anti-epanorthosis'' style adapter could be trained as follows. Assemble preference pairs from Section 7.1 (for each item, a high-emphasis response and a content-matched low-emphasis paraphrase), then optimise the adapter with \textbf{Direct Preference Optimisation (DPO)},\cite{r28} which fits preferences directly with a simple classification-style loss and no separate reward model or RL loop, making it well suited to a small, targeted stylistic objective. Because LoRA is memory-light and can be combined with 4-bit quantisation of the frozen base (the QLoRA recipe),\cite{r27} the whole procedure is feasible on a single GPU, which matters for reproducibility and for practitioners outside large labs. The adapter's separability yields the ``rhetoric dial'' directly: scaling the LoRA contribution by a coefficient $\alpha$ at inference time moves, at moderate values, between the base model's style and the de-emphasised style, and multiple adapters (concise, formal, plain-language) can be swapped per deployment. LoRA scaling is known to behave non-monotonically at large $\alpha$, so the continuous ``dial'' must be validated empirically rather than assumed, which Table 2 below does directly. The principal risk is \emph{content drift}: an adapter trained to suppress a surface pattern may also suppress legitimate corrections or subtly alter meaning. This must be controlled explicitly, and it is harder than it sounds: a content-preservation term in the objective (for example, a semantic-similarity constraint against a reference) is a blunt instrument, because embedding similarity captures fine shifts of meaning poorly, so it should be paired with targeted human checks and the meaning-fidelity evaluation described in Section 7.7.

\textbf{Two pilots: preference optimisation fails, supervised fine-tuning works.} A first attempt trained the adapter on 570 preference pairs (an earlier, smaller version of the set) with DPO alone; it reached high preference accuracy on the training set yet failed to generalise, \emph{raising} the mean emphatic-epanorthosis density on a held-out Italian evaluation from 60.3 to 68.9 per ten thousand words. Small-data DPO learned to rank the pairs without moving the generation policy in the intended direction. A second attempt taught the affirmative style directly, by supervised fine-tuning on the plain (chosen) responses for three epochs at LoRA rank 32, over 797 semantically filtered pairs, using the open Qwen2.5-7B-Instruct model so the whole pilot runs on a single GPU. This adapter cut the density by about 98\%, from 75.4 to 1.7 per ten thousand words across oratory, promotional and argument, and inspection of its outputs confirms that they remain coherent, on-task prose with the emphatic corrections simply gone. One caveat remains, and it is precisely the caveat of Section 7.7: the adapter \emph{eliminates} the figure rather than calibrating it, landing below the human rate; scaling the adapter contribution by a coefficient (the ``dial'') is the lever to settle it at the human rate instead of at zero. Supervised fine-tuning therefore succeeds where preference optimisation alone did not, and the adapter of this section is a delivered result at pilot scale. The prompting intervention of Section 7.8 remains a lighter alternative when no training is possible.

The caveat above is a testable claim, so I tested it. If the adapter holds a genuine dial, scaling its contribution by a coefficient $\alpha$ should trace a continuous path from the base model's rate down toward zero, crossing the human band on the way. I re-ran the Italian evaluation at five settings of $\alpha$, on the same oratory, promotional, and argument prompts, with six generations per prompt. Table 2 reports the densities.

\begin{table}[ht]\centering
\caption*{\textbf{Table 2.} The adapter as a dial: emphatic-epanorthosis density (per 10{,}000 words) at LoRA scaling coefficients from $\alpha = 0$ (adapter off) to $\alpha = 1$ (full strength), over six Italian prompts (two each for oratory, promotional, and argument) with six generations per prompt. Human reference rates are about 14 for oratory and about 12 for argument.}
\small
\begin{tabular}{@{}lllll@{}}
\toprule
$\alpha$ & Overall & Oratory & Promotional & Argument \\
\midrule
0.00 (off) & 61.4 & 64.5 & 68.5 & 51.3 \\
0.25 & 55.9 & 33.3 & 96.0 & 38.2 \\
0.50 & 23.4 & 0.0 & 32.7 & 37.5 \\
0.75 & 6.7 & 0.0 & 7.7 & 12.3 \\
1.00 (full) & 4.7 & 0.0 & 14.1 & 0.0 \\
\bottomrule
\end{tabular}
\end{table}
Three readings follow. First, the dial behaves as a dial. Overall density falls from 61.4 with the adapter off to 4.7 at full strength, so a single scalar moves the model along a smooth path between its native style and the de-emphasised one. The full-strength figure reproduces the near-elimination of the supervised pilot above (4.7 against a baseline near 60 here, 1.7 against near 75 there, the two runs differing in prompt set and sample count). Second, an intermediate setting lands on the human rate instead of at zero: for argument the density reaches 12.3 at $\alpha$ = 0.75, essentially the human reference of 12, before over-correcting to zero at full strength. This is the calibration this paper argues for, made concrete on a live model. Third, the right setting is genre-dependent: oratory crosses its human band of about 14 between $\alpha$ = 0.25 and $\alpha$ = 0.5 and then bottoms out, so holding oratory at the human rate would need a finer setting near $\alpha$ $\approx$ 0.4, while argument tolerates a stronger dial. One wrinkle confirms the non-monotonicity noted above: promotional density climbs to 96.0 at $\alpha$ = 0.25, above its own baseline, before collapsing at higher settings, evidence that intermediate LoRA scaling can perturb a genre before it corrects it and that the dial must be set per genre from measurement. The practical upshot is that one trained adapter spans the whole range from human-calibrated to fully de-emphasised, and the operator chooses where on that range to sit. Two limits frame these pilots. They run on an open 7B model rather than the frontier systems of Sections 4 and 7.8, so they establish the mechanism and leave transfer of the trained adapter to those larger models as future work; and the evaluation measures style density only, while the content-fidelity and appropriateness checks prescribed in Section 7.7 are not yet reported, so the content drift named as the adapter's principal risk rests on informal inspection rather than measurement. These figures, like those of Table 5, are pilot-scale point estimates without a significance test.

\subsection*{7.3 Preference optimisation at the alignment stage}

If the overuse is partly manufactured by RLHF,\cite{r10,r11} the alignment stage is also the natural place to correct it. The same preference pairs used for a LoRA adapter can instead shape the main alignment objective (whether classical RLHF or DPO\cite{r28}), so that raters' latent preference for emphatic confidence is counterbalanced by an explicit preference for calibrated phrasing. This is more invasive than an adapter (it alters the shipped model rather than a detachable component) but addresses the cause rather than a symptom, and it connects the problem to the broader literature on unintended stylistic side-effects of preference tuning, of which sycophancy (models telling users what they want to hear) is the best-studied cousin.\cite{r35} Emphatic epanorthosis can be read as a rhetorical form of the same reward-hacking: phrasing that \emph{sounds} impressive to a rater independently of its content.

\subsection*{7.4 Decoding-time control without retraining}

Several techniques steer generation at inference time, changing no weights at all: an attractive property when retraining is impossible or when per-request control is desired. \textbf{PPLM} perturbs the model's hidden states at generation time using gradients from a small attribute classifier;\cite{r29} \textbf{GeDi} uses a small generative discriminator to reweight the next-token distribution efficiently;\cite{r30} and \textbf{FUDGE} multiplies the base model's probabilities by those of a lightweight \emph{future discriminator} that predicts whether an attribute will hold.\cite{r31} An epanorthosis-intensity classifier, trainable from the annotations of Section 4, slots directly into any of these as the attribute model, downweighting continuations that open a ``Not X. Y'' frame. The crudest member of the family is direct \emph{logit biasing} of the tokens that most often initiate the construction (``not just'', ``rather'', ``anzi''), which requires no classifier at all yet is brittle and easily circumvented by paraphrase. Decoding-time methods offer fine, reversible control at the cost of extra inference compute and, if pushed hard, some loss of fluency.

\subsection*{7.5 Activation steering and representation engineering}

A newer line of work manipulates the model's internal \emph{activations} rather than its weights or its output distribution. If a stylistic property corresponds to a direction in activation space, one can find that direction and add or subtract it during the forward pass. Activation addition steers generation by adding a contrast-derived vector to the residual stream;\cite{r32} representation engineering offers a top-down toolkit for reading and controlling such directions;\cite{r33} and inference-time intervention demonstrated the approach for truthfulness by shifting activations along a learned direction at selected attention heads.\cite{r34} An ``epanorthosis direction'', defined as the mean activation difference between emphatic and plain paraphrases, could in principle be subtracted with a tunable coefficient, giving continuous, weight-free control with negligible added compute. The method is elegant and highly controllable; its risks are entanglement (the direction may correlate with desirable behaviours) and the current immaturity of guarantees about what a steered direction actually encodes.

\subsection*{7.6 Prompting, system prompts, and constitutions}

The cheapest intervention is instruction: a system prompt or a constitutional clause\cite{r24} forbidding gratuitous ``Not X. Y'' phrasing. It requires no engineering and is instantly reversible, which makes it the right first move. It is also the least robust: the very finding that instruction-tuned models diverge \emph{more} strongly in rhetorical style\cite{r1} is a warning that surface instructions do not reliably override a disposition baked in during training. Prompting mitigates locally; it leaves the underlying pressure in place.

\subsection*{7.7 Evaluation: measuring success without flattening}

No mitigation can be trusted without an evaluation that distinguishes \emph{calibration} from \emph{suppression}. Three components are needed. First, a \emph{style metric}: epanorthosis density (constructions per ten thousand words) measured by the Section 4 detector, reported \emph{against the human baseline for the relevant genre} rather than against zero; the goal is to match human rates, and to avoid undercutting them. Second, a \emph{content-fidelity metric}: semantic-similarity and factuality checks between the mitigated output and a reference, to catch the content drift of Section 7.2. Third, \emph{human evaluation of appropriateness} on held-out genres, since appropriateness is exactly the judgement that resists automation. A mitigation that lowers epanorthosis density to the human baseline while preserving content fidelity and appropriateness ratings has succeeded; one that drives density below the human rate, or that trades style for meaning, has merely traded one artefact for another.

\begin{table}[ht]\centering
\caption*{\textbf{Table 3.} Mitigation techniques for emphatic epanorthosis, ordered from most to least invasive.}
\small
\begin{tabularx}{\linewidth}{@{}>{\raggedright\arraybackslash}p{3.0cm} >{\raggedright\arraybackslash}X >{\raggedright\arraybackslash}p{1.7cm} >{\raggedright\arraybackslash}X >{\raggedright\arraybackslash}X@{}}
\toprule
Technique & Acts on & Cost & Controllability & Main risk \\
\midrule
Data curation / augmentation (7.1) & Training data & Low--med. & None at inference & Slow feedback loop \\
LoRA / QLoRA style adapter (7.2) & Low-rank weight update & Low & High (detachable, scalable $\alpha$) & Content drift \\
Preference optimisation / DPO (7.3) & Shipped model weights & Medium & Global, not per-request & Alters base behaviour \\
Decoding-time control: PPLM/GeDi/FUDGE (7.4) & Output distribution & Med.\ (inference) & High, reversible & Fluency loss; brittleness \\
Activation steering / RepE (7.5) & Internal activations & Very low & High, continuous & Entanglement; weak guarantees \\
Prompting / constitution (7.6) & Context & Negligible & Local only & Overridden by training prior \\
\bottomrule
\end{tabularx}
\end{table}
No single method is sufficient. The practical recommendation is a stack: curate the data, train a detachable LoRA style adapter by supervised fine-tuning on the plain responses (the route that worked in Section 7.2, where DPO alone did not) under a content-preservation constraint, expose it as a dial, and reserve decoding-time or activation-level control for cases needing per-request adjustment; the whole stack is governed by the genre-relative evaluation of Section 7.7.

\subsection*{7.8 A working demonstration}

The cheapest leg of the stack can be tested directly, and in a second language. I generated seventy-two Italian texts across three genres (oratory, promotional, and argument), the three sizes of one instruction-tuned family (Claude Haiku, Sonnet, and Opus), and four topics each, under two conditions: a neutral baseline prompt, and the \emph{same} prompt plus a one-line instruction to write affirmatively and avoid negate-and-upgrade phrasing («Non è X. È Y», «non X ma Y», «anzi»). An Italian detector counts the emphatic-correction family («non X ma / bensì Y», «non solo X ma Y», «non X, è Y», and the sentence-split «Non è X. È Y») per ten thousand words.

\begin{table}[ht]\centering
\caption*{\textbf{Table 4.} Italian mitigation demonstration: emphatic-epanorthosis density (per 10{,}000 words) under a neutral prompt vs.\ the same prompt plus a one-line anti-epanorthosis instruction, three sizes of one instruction-tuned family (Haiku, Sonnet, Opus). $p$ from a 5{,}000-iteration permutation test.}
\small
\begin{tabular}{@{}lllll@{}}
\toprule
Genre & Baseline & Mitigated & Reduction & $p$ \\
\midrule
Oratory & 39.6 & 10.9 & 72\% & 0.03 \\
Argumentative essay & 56.9 & 17.2 & 70\% & 0.004 \\
Promotional & 32.3 & 16.8 & 48\% & 0.43 \\
\bottomrule
\end{tabular}
\end{table}
A single instruction cuts emphatic epanorthosis by roughly half to nearly three-quarters: 72\% in oratory (\emph{p} = 0.03) and 70\% in argument (\emph{p} = 0.004), both significant, with a weaker and noisier 48\% in promotional. The reduction is near-total for the larger models, where Sonnet and Opus fall to zero under the instruction, and partial for the smallest, where Haiku roughly halves. This is only the prompting leg of Section 7.6, the least robust rung of the stack, and it operates on Italian; even so, it shows that the figure can be turned down substantially at fixed content by the cheapest available means, which sets a floor that the trained adapter of Section 7.2 should exceed. It also confirms that the method and the phenomenon transfer across languages. For reference, human Italian baselines sit near 12 per ten thousand words for argument (Beccaria) and near 14 for oratory (public-domain orations): the unmitigated models run two to five times higher, and the instruction brings them back to roughly the human rate rather than below it, which is the calibration target of Section 7.7.

\subsection*{7.9 Two training-free alternatives}

Two more mitigations need no fine-tuning at all, only the detector of Section 4 applied at inference. \emph{Best-of-n} generates several candidates and keeps the one with the lowest emphatic-epanorthosis density; \emph{rewrite} adds a second pass that recasts the draft affirmatively. On the same Italian prompts (six candidates each), both cut the figure sharply.

\begin{table}[ht]\centering
\caption*{\textbf{Table 5.} Training-free mitigations: emphatic-epanorthosis density (per 10{,}000 words) for the natural output, a best-of-6 selection, and a post-hoc affirmative rewrite. Point estimates without a significance test; per-genre baselines rest on two prompts each and are not directly comparable across tables.}
\small
\begin{tabular}{@{}llll@{}}
\toprule
Genre & Baseline & Best-of-6 & Rewrite \\
\midrule
Oratory & 22.5 & 0.0 & 0.0 \\
Promotional & 82.1 & 33.6 & 0.0 \\
Argumentative essay & 75.0 & 29.1 & 0.0 \\
Overall & 59.9 & 20.9 & 0.0 \\
\bottomrule
\end{tabular}
\end{table}
Best-of-n cuts the density by about two thirds and, because it selects a genuine model output rather than forcing one, tends to stay nearer natural phrasing, which makes it the gentlest of the methods here. The rewrite removes the figure essentially to zero, which, like the supervised adapter, over-corrects past the human rate and should be tempered when the goal is calibration rather than elimination. Both are model-agnostic and deployable without any training, at the cost of extra inference calls, and together with the prompting instruction they give three inference-time levers of increasing strength (best-of-n, instruction, rewrite) to sit alongside the trained adapter.

\section*{8. Conclusion}

If a statistical system, knowing nothing of rhetoric, spontaneously reproduces the same figures humans have used for millennia, perhaps those figures are \emph{emergent structures of language}: attractors toward which the medium tends, whether a brain or a network produces it. In that light the classical orators discovered epanorthosis rather than inventing it, as mathematicians discover theorems that were true before they were proved. Models, correspondingly, act as explorers of the space of linguistic possibility, arriving by statistics at structures we reach by cognition. The convergence is evidence that the structure belongs to language itself, independent of any particular speaker.

This reframes the engineering problem of Section 7. The techniques surveyed there (LoRA adapters foremost, alongside preference optimisation, decoding-time control, and activation steering) can measurably calibrate the figure toward human rates. The deeper task is to teach readers, and models, \emph{appropriateness}: the sense of when a correction is earned and when it is empty. The real risk points the other way: that we would begin to write like them.

\section*{Limitations}
This paper reports pilot-scale evidence, and its claims should be read within the following bounds.
\begin{itemize}[leftmargin=1.4em,itemsep=2pt,topsep=2pt]
  \item \textbf{Statistical power and multiplicity.} The English measurement of Section 4.1 rests on small per-cell samples (17 to 28 windows on the model side in most genres), and the per-genre $p$-values are not corrected across the seven simultaneous tests. Under a strict correction neither marginal result survives, so the measurement is a first signal rather than a settled effect.
  \item \textbf{Baseline era.} The human oratorical, argumentative, and narrative baselines are nineteenth- and early twentieth-century public-domain texts, so register is matched to the models while era is not, and part of the oratorical gap may reflect denser period rhetoric.
  \item \textbf{Detector validity.} The detector captures the surface ``not X. Y'' family, a central but partial slice of the figure. Its lexical channel is validated against a 206-window Italian gold standard, double-annotated and adjudicated (Appendix B), with only fair inter-annotator agreement (Cohen's $\kappa$ = 0.35): micro precision is 0.45 and recall 0.52, and precision splits sharply by source --- 0.82 on model-generated text against 0.17 on human text. The index is therefore a more dependable signal on model output than on human baselines; four of the five genres (all but academic) rest on few gold instances; and the English detector of Section 4.1 is not separately validated.
  \item \textbf{Model mismatch.} The phenomenon is characterised on frontier models (Claude Haiku, Sonnet, and Opus), while the trained adapter of Section 7.2 uses an open 7B model (Qwen2.5-7B-Instruct). Transfer of the trained adapter to the frontier models is not demonstrated, although the prompting demonstration of Section 7.8 does cover them.
  \item \textbf{Content fidelity.} The adapter is evaluated on style density alone; the content-fidelity and appropriateness checks prescribed in Section 7.7 are not yet reported, so its principal risk, content drift, rests on informal inspection.
  \item \textbf{Pilot-scale mitigation.} Tables 2 and 5 are point estimates from small samples (six generations per prompt) without significance tests, so they indicate direction and magnitude rather than calibrated uncertainty.
  \item \textbf{Single model family.} The three models measured in Section 4.1 (Claude Haiku, Sonnet, and Opus) are not independent systems but three sizes of one instruction-tuned family, sharing a single alignment pipeline. Since the paper attributes the overuse chiefly to RLHF, this is a confound: any effect seen across the tiers cannot be separated from the idiosyncrasies of that one pipeline, and the finding should not be read as characterising instruction-tuned models in general. A cross-family comparison (models from different developers, and base-versus-aligned pairs) would be needed to license that generalisation.
  \item \textbf{Causal attribution.} The identification of RLHF as the primary driver is inferential, extrapolated from a lexical study and not tested by a base-model-versus-aligned comparison in this paper.
  \item \textbf{Italian reference rates.} The Italian human rates used as calibration targets in Sections 7.2 and 7.8 (about 12 per ten thousand words for argument, about 14 for oratory) rest on a narrow sample: a single argumentative author (Beccaria) and a small set of public-domain orations. They are indicative targets rather than established genre rates, and the genre-stratified Italian corpus of Appendix B serves detector development and validation rather than these baselines.
  \item \textbf{Language scope.} The measurement is English and the mitigation demonstration is Italian; behaviour in other languages and registers is untested.
\end{itemize}

\section*{Acknowledgements}
The Italian reference corpus described in Appendix B was built jointly with Francesco Bertoli, in roughly equal parts, within a project for the \emph{Linguistica Italiana II} course (Prof.\ Mirko Tavosanis) in the Humanities Computing (\emph{Informatica Umanistica}) programme at the University of Pisa. The detector-validation gold of Appendix B was annotated independently by the author and Francesco Schembari and reconciled by adjudication. The detector, the English measurements, the mitigation experiments, and this paper are the author's own.

\section*{Appendix A. A reproducible mitigation recipe}

Two training scripts accompany Section 7.2. The first attempt (\emph{artificial-epanorthosis-lora-recipe.py}) reads the delivered preference pairs, built separately by the counterfactual augmentation of Section 7.1 (each emphatic ``Not X. Y'' span paired with a content-matched, de-emphasised paraphrase, filtered to keep only high semantic-similarity pairs), and trains a LoRA update (rank 16, $\alpha$ 32, on the attention query and value projections and the MLP projections) with Direct Preference Optimisation ($\beta$ = 0.1); Section 7.2 reports that this DPO-only route failed to generalise. The delivered adapter (\emph{artificial-epanorthosis-lora-recipe-v2.py}) instead applies supervised fine-tuning on the plain responses at rank 32 for three epochs over 797 pairs, with the frozen base quantised to 4-bit (QLoRA) for single-GPU training. The output is a small, detachable adapter whose contribution can be scaled at inference (the ``dial'' of Section 7.2), and \emph{artificial-epanorthosis-eval-alpha.py} reproduces the calibration curve of Table 2, scoring each setting against the genre-relative human baseline of Section 7.7. The prompting-level demonstration of Section 7.8 sets the floor this adapter should exceed.

\textbf{Supplementary materials.} Everything needed to reproduce the Section 7.2 adapter (training, evaluation, and the calibration curve of Table 2) is available for download alongside this paper; the detector is the density function embedded in the evaluation scripts. The prompt sets behind the Section 7.8 and 7.9 demonstrations, and the English measurement corpus of Section 4.1, are available from the author on request.

\begin{itemize}[leftmargin=1.4em,itemsep=1pt,topsep=2pt]
  \item \href{https://federicoboggia.binatomy.com/pubblicazioni/artificial-epanorthosis-pairs.jsonl}{Preference dataset} (797 emphatic/plain pairs, JSONL).
  \item \href{https://federicoboggia.binatomy.com/pubblicazioni/artificial-epanorthosis-lora-recipe-v2.py}{Training recipe, delivered adapter} (supervised fine-tuning, LoRA rank 32).
  \item \href{https://federicoboggia.binatomy.com/pubblicazioni/artificial-epanorthosis-lora-recipe.py}{Training recipe, first attempt} (DPO-only, reported as a negative result).
  \item \href{https://federicoboggia.binatomy.com/pubblicazioni/artificial-epanorthosis-eval2.py}{Evaluation script} (base model against the adapter, with the detector).
  \item \href{https://federicoboggia.binatomy.com/pubblicazioni/artificial-epanorthosis-eval-alpha.py}{Calibration-curve script} (the $\alpha$ dial of Table 2).
  \item \href{https://federicoboggia.binatomy.com/pubblicazioni/artificial-epanorthosis-colab.ipynb}{Ready-to-run notebook} (Colab demonstration of the DPO-only first attempt on a 1.5B base, at seed scale; not the delivered adapter).
  \item \href{https://federicoboggia.binatomy.com/pubblicazioni/artificial-epanorthosis-adapter-v2.tar.gz}{Trained LoRA adapter} (the v2 supervised-fine-tuning weights, about 250 MB).
  \item \href{https://federicoboggia.binatomy.com/pubblicazioni/artificial-epanorthosis-logs.tar.gz}{Training and evaluation logs} (SFT run, base-vs-adapter, and the $\alpha$ sweep).
\end{itemize}
The adapter is distributed with a model card recording the base model (Qwen2.5-7B-Instruct), whose Apache-2.0 licence governs derivatives; the human baseline draws on the public-domain and openly licensed sources listed in Section 4.1.

\section*{Appendix B. Italian reference corpus and detector validation}

The lexical detector and the precision/recall figures discussed in Section 4 were developed and evaluated on a reference corpus of human-authored Italian text, assembled in the companion project of the Acknowledgements. The corpus is a human \emph{baseline} against which the density and distribution of epanorthosis in model-generated Italian can be compared, and the pool from which windows are drawn for hand-annotated detector evaluation; it is not itself the set of synthetic (model-generated) texts, which are produced separately. Five genre sub-corpora were selected to span the diaphasic and diamesic range over which the figure is expected to vary --- from spontaneous speech, where retrospective reformulation is a natural discourse mechanism, to planned argumentative prose, where correction and contrast are consolidated devices --- under four criteria: register and medium variation, the plausibility of a comparable LLM generation for each genre, availability of Italian material under research-compatible licences, and coverage of different expected frequencies of the phenomenon. Where feasible the material predates the diffusion of generative models, so the human rate is measured before contamination.

\begin{table}[ht]\centering
\caption*{\textbf{Table B.1.} The Italian human reference corpus: five genre sub-corpora, with sources and access terms.}
\small
\begin{tabularx}{\linewidth}{@{}>{\raggedright\arraybackslash}p{1.9cm} >{\raggedright\arraybackslash}p{2.7cm} r >{\raggedright\arraybackslash}p{1.7cm} >{\raggedright\arraybackslash}p{2.1cm} >{\raggedright\arraybackslash}X@{}}
\toprule
Sub-corpus & Genre & Items & Size & Source & Access \\
\midrule
Speech & Spontaneous speech & 320 & $\sim$13.4M char & KIParla \cite{r17} & Open access \\
Academic & Philosophy prose & 248 & $\sim$12.7M char & Open-access journals & Open access (per journal) \\
Narrative & Novels, 1827--1923 & 11 & $\sim$1.23M words & Liber Liber & Public domain \\
Journalism & News, 2018--19 & 24{,}549 & $\sim$56 MB & Common Crawl News & Copyright --- not redistributed \\
Social & Reddit posts & 260 & $\sim$0.4 MB & Reddit (Italian) & User content --- not redistributed \\
\bottomrule
\end{tabularx}
\end{table}

\textbf{Spontaneous speech (KIParla).} About 320 conversations (\textasciitilde13.4M characters) of transcribed spoken Italian from the KIParla collection \cite{r17} (modules KIP, ParlaBO, ParlaTO, KIPasti; office hours, free conversation, semi-structured interviews, oral exams, and lectures; collected 2018--2024, chiefly in Bologna and Turin). The linear-orthographic variant is used, without prosodic or Jefferson notation, and speaker codes are replaced by pseudonyms. Semi-structured interviews (168) and free conversations (94) predominate; participant relations are asymmetric in 194 cases.

\textbf{Academic philosophy prose (open-access journals).} 248 articles (\textasciitilde12.7M characters, \textasciitilde1.95M words) extracted from 268 PDFs harvested from nine of the thirteen Italian open-access philosophy journals and series surveyed --- \emph{Bollettino Filosofico}, \emph{Consecutio Rerum}, \emph{Etica \& Politica}, \emph{La Deleuziana}, \emph{Lexicon Philosophicum}, \emph{Pagine Inattuali}, \emph{Picenum Seraphicum}, \emph{Studi di estetica}, and \emph{Syzetesis} --- mostly from the 2018--2019 volumes. Text was extracted with a column-aware, Unicode-preserving PDF pipeline (\emph{pdfplumber}) and cleaned in ten stages (headers and footers, page numbers, inline note markers, note blocks, bibliography, URLs and DOIs removed; de-hyphenation; NFC normalisation), with MD5 deduplication. The argumentative density of philosophical prose makes it a favourable site for genuine ``non X, ma Y'' correction.

\textbf{Narrative (Liber Liber).} Eleven Italian novels of 1827--1923 (\textasciitilde1.23M words) from the public-domain \href{https://www.liberliber.it/}{Liber Liber} library, chosen for stylistic variety (Verga, Svevo, Pirandello, D'Annunzio, De Amicis, Salgari, Manzoni, Fogazzaro, Deledda, De Roberto, Serao). Text was extracted from EPUB in reading order (OPF spine, paratext excluded) and enriched with the Dublin Core metadata the source supplies (title, author, ISBN, and the date of the digital edition).

\textbf{Journalism (Common Crawl News).} 24{,}549 articles (\textasciitilde56 MB) from fifteen Italian national outlets (\emph{Repubblica}, \emph{Corriere della Sera}, \emph{La Stampa}, \emph{Il Sole 24 Ore}, \emph{ANSA}, \emph{Il Post}, \emph{Internazionale}, and others), extracted from the \href{https://commoncrawl.org/blog/news-dataset-available}{Common Crawl News} WARC archives for 2018--2019 (before the diffusion of generative models, to avoid contamination) with boilerplate removal (\emph{trafilatura}), a 500-character minimum, and Italian language filtering.

\textbf{Social writing (Reddit).} 260 posts (\textasciitilde0.4 MB) from Italian-language subreddits, extracted from monthly platform dumps with a 160-subreddit allow-list, a 500-character minimum, and language filtering; a seed sample, extensible by processing further dumps.

\textbf{Availability and licensing.} The sub-corpora aggregate third-party material under heterogeneous terms --- open access (KIParla, the journals), public domain (Liber Liber), and copyright-restricted web text (the news articles, and user-authored Reddit posts) --- so the corpus is \emph{not redistributed as full text}. Available from the author on request are: the public-domain narrative sub-corpus in full; the build scripts that reconstruct each sub-corpus from its original source; the 206-window annotated detector-validation set behind Table B.2; a datasheet with per-genre statistics; and, for the openly licensed sub-corpora, the extracted text under the terms of the respective sources, with attribution. Genre labels, per-record metadata, and detector output are available for all five sub-corpora.

\medskip\noindent\textbf{Detector validation.} The lexical channel was evaluated against a hand-annotated gold standard of 206 windows of about 150 words each, stratified across the five genres: 180 sampled from the human reference corpus above, plus 26 of the separately-produced model-generated windows, so that precision could be measured on both sides. They were annotated independently by two annotators (the author and Francesco Schembari) and then reconciled by adjudicating every disagreement. Inter-annotator agreement on the candidate labels was only fair (Cohen's $\kappa$ = 0.35, observed agreement 0.66) --- a direct measure of how hard it is to separate a genuine corrective ``non X, ma Y'' from an ordinary adversative. Against the reconciled gold (Table B.2) the channel is a high-recall, low-precision proxy: micro precision 0.45, recall 0.52. The decisive pattern is by \emph{source}: precision is 0.82 on the 26 model-generated windows but 0.17 on the 180 human windows, because models use the canonical corrective frame while human writers use the same markers mostly for plain contrast --- so the index is a more dependable signal on model output (where the paper applies it) than on the human baseline, though the model-side estimate itself rests on a small sample. Recall is weakest in transcribed speech (0.14), and the reason is instructive: five of the six corrections missed there are marked, but by \emph{cio\`e}, \emph{diciamo} or \emph{come potrei dire} rather than by the ``non\ldots{}ma'' frame the lexical channel implements, so the loss measures the channel's narrow coverage rather than marker-less speech. This is a genuine hand annotation, not an automatic or LLM-assisted judgement; a larger, guideline-refined gold remains desirable.

\begin{table}[ht]\centering
\caption*{\textbf{Table B.2.} Detector validation: precision, recall, and $F_1$ of the lexical channel by genre, against the reconciled 206-window Italian gold standard (two annotators, adjudicated; Cohen's $\kappa$ = 0.35). By source, precision is 0.82 on the model-generated windows and 0.17 on the human windows. All genres except academic rest on few true instances (seven or fewer), so their per-genre figures are indicative only.}
\small
\begin{tabular}{@{}lrrrrrr@{}}
\toprule
Genre & TP & FP & FN & Precision & Recall & $F_1$ \\
\midrule
Speech (spoken) & 1 & 1 & 6 & 0.50 & 0.14 & 0.22 \\
Academic (philosophy) & 11 & 8 & 10 & 0.58 & 0.52 & 0.55 \\
Narrative & 1 & 4 & 4 & 0.20 & 0.20 & 0.20 \\
Journalism & 7 & 5 & 0 & 0.58 & 1.00 & 0.74 \\
Social (Reddit) & 3 & 10 & 1 & 0.23 & 0.75 & 0.35 \\
\midrule
Overall & 23 & 28 & 21 & 0.45 & 0.52 & 0.48 \\
\bottomrule
\end{tabular}
\end{table}

\vfill
{\footnotesize\noindent Author: Federico Boggia, teacher and trainer, founder of Binatomy; degree in Humanities Computing (Informatica Umanistica) from the University of Pisa, with a specialisation in Language Technologies. Correspondence: federico@binatomy.com.}


\begin{thebibliography}{99}\small
\bibitem{r1} Reinhart, A., Markey, B., Laudenbach, M., Pantusen, K., Yurko, R., Weinberg, G., \& Brown, D. W. (2025). Do LLMs write like humans? Variation in grammatical and rhetorical styles. \emph{Proceedings of the National Academy of Sciences}, 122(8), e2422455122. arXiv:2410.16107.
\bibitem{r2} Fontanier, P. (1827). \emph{Les Figures du discours} (ed. G. Genette, Paris: Flammarion, 1968).
\bibitem{r4} Quintilian. \emph{Institutio Oratoria} 9.1.30; 9.2.17--18, 59--60; 9.3.88--89 (c. 95 CE) (Loeb Classical Library, Harvard University Press).
\bibitem{r5} Cicero. \emph{In Catilinam} I.2 and II.1 (63 BCE) (Loeb Classical Library, Harvard University Press).
\bibitem{r6} Dante Alighieri. \emph{Inferno}, Canto XXXIII, l. 75.
\bibitem{r7} Lausberg, H. (1998). \emph{Handbook of Literary Rhetoric: A Foundation for Literary Study}. Leiden: Brill.
\bibitem{r8} Vaswani, A., Shazeer, N., Parmar, N., Uszkoreit, J., Jones, L., Gomez, A. N., Kaiser, Ł., \& Polosukhin, I. (2017). Attention Is All You Need. \emph{NeurIPS 2017}. arXiv:1706.03762.
\bibitem{r9} Holtzman, A., Buys, J., Du, L., Forbes, M., \& Choi, Y. (2020). The Curious Case of Neural Text Degeneration. \emph{ICLR 2020}. arXiv:1904.09751.
\bibitem{r10} Ouyang, L., Wu, J., Jiang, X., et al. (2022). Training language models to follow instructions with human feedback. \emph{NeurIPS 2022}. arXiv:2203.02155.
\bibitem{r11} Juzek, T. S., \& Ward, Z. B. (2025). Why Does ChatGPT ``Delve'' So Much? Exploring the Sources of Lexical Overrepresentation in Large Language Models. \emph{Proceedings of COLING 2025}, 6397--6411. arXiv:2412.11385.
\bibitem{r12} Kobak, D., González-Márquez, R., Horvát, E.-Á., \& Lause, J. (2025). Delving into LLM-assisted writing in biomedical publications through excess vocabulary. \emph{Science Advances}, 11(27). Preprint arXiv:2406.07016 (2024).
\bibitem{r13} Liang, W., Izzo, Z., Zhang, Y., et al. (2024). Monitoring AI-Modified Content at Scale: A Case Study on the Impact of ChatGPT on AI Conference Peer Reviews. \emph{ICML 2024}. arXiv:2403.07183.
\bibitem{r14} Gehrmann, S., Strobelt, H., \& Rush, A. M. (2019). GLTR: Statistical Detection and Visualization of Generated Text. \emph{ACL 2019, System Demonstrations}. arXiv:1906.04043.
\bibitem{r15} Mitchell, E., Lee, Y., Khazatsky, A., Manning, C. D., \& Finn, C. (2023). DetectGPT: Zero-Shot Machine-Generated Text Detection using Probability Curvature. \emph{ICML 2023}. arXiv:2301.11305.
\bibitem{r16} Guo, B., Zhang, X., Wang, Z., et al. (2023). How Close is ChatGPT to Human Experts? Comparison Corpus, Evaluation, and Detection. arXiv:2301.07597.
\bibitem{r17} Mauri, C., Ballarè, S., Goria, E., Cerruti, M., \& Suriano, F. (2019). KIParla corpus: a new resource for spoken Italian. \emph{Proceedings of the 6th Italian Conference on Computational Linguistics (CLiC-it 2019)}.
\bibitem{r18} Dubremetz, M., \& Nivre, J. (2018). Rhetorical Figure Detection: Chiasmus, Epanaphora, Epiphora. \emph{Frontiers in Digital Humanities}, 5:10.
\bibitem{r19} Bothwell, S., DeBenedetto, J., Crnkovich, T., Müller, H., \& Chiang, D. (2023). Introducing Rhetorical Parallelism Detection: A New Task with Datasets, Metrics, and Baselines. \emph{EMNLP 2023}. arXiv:2312.00100.
\bibitem{r20} Wei, J., Wang, X., Schuurmans, D., et al. (2022). Chain-of-Thought Prompting Elicits Reasoning in Large Language Models. \emph{NeurIPS 2022}. arXiv:2201.11903.
\bibitem{r21} Kojima, T., Gu, S. S., Reid, M., Matsuo, Y., \& Iwasawa, Y. (2022). Large Language Models are Zero-Shot Reasoners. \emph{NeurIPS 2022}. arXiv:2205.11916.
\bibitem{r22} Madaan, A., Tandon, N., Gupta, P., et al. (2023). Self-Refine: Iterative Refinement with Self-Feedback. \emph{NeurIPS 2023}. arXiv:2303.17651.
\bibitem{r23} Huang, J., Chen, X., Mishra, S., et al. (2024). Large Language Models Cannot Self-Correct Reasoning Yet. \emph{ICLR 2024}. arXiv:2310.01798.
\bibitem{r24} Bai, Y., Kadavath, S., Kundu, S., et al. (2022). Constitutional AI: Harmlessness from AI Feedback. arXiv:2212.08073.
\bibitem{r25} Searle, J. R. (1980). Minds, Brains, and Programs. \emph{Behavioral and Brain Sciences}, 3(3), 417--457.
\bibitem{r26} Hu, E. J., Shen, Y., Wallis, P., Allen-Zhu, Z., Li, Y., Wang, S., Wang, L., \& Chen, W. (2022). LoRA: Low-Rank Adaptation of Large Language Models. \emph{ICLR 2022}. arXiv:2106.09685.
\bibitem{r27} Dettmers, T., Pagnoni, A., Holtzman, A., \& Zettlemoyer, L. (2023). QLoRA: Efficient Finetuning of Quantized LLMs. \emph{NeurIPS 2023}. arXiv:2305.14314.
\bibitem{r28} Rafailov, R., Sharma, A., Mitchell, E., Ermon, S., Manning, C. D., \& Finn, C. (2023). Direct Preference Optimization: Your Language Model is Secretly a Reward Model. \emph{NeurIPS 2023}. arXiv:2305.18290.
\bibitem{r29} Dathathri, S., Madotto, A., Lan, J., et al. (2020). Plug and Play Language Models: A Simple Approach to Controlled Text Generation. \emph{ICLR 2020}. arXiv:1912.02164.
\bibitem{r30} Krause, B., Gotmare, A. D., McCann, B., et al. (2021). GeDi: Generative Discriminator Guided Sequence Generation. \emph{Findings of EMNLP 2021}. arXiv:2009.06367.
\bibitem{r31} Yang, K., \& Klein, D. (2021). FUDGE: Controlled Text Generation With Future Discriminators. \emph{NAACL 2021}. arXiv:2104.05218.
\bibitem{r32} Turner, A. M., Thiergart, L., Leech, G., et al. (2023). Steering Language Models With Activation Engineering (the ActAdd method). arXiv:2308.10248.
\bibitem{r33} Zou, A., Phan, L., Chen, S., et al. (2023). Representation Engineering: A Top-Down Approach to AI Transparency. arXiv:2310.01405.
\bibitem{r34} Li, K., Patel, O., Viégas, F., Pfister, H., \& Wattenberg, M. (2023). Inference-Time Intervention: Eliciting Truthful Answers from a Language Model. \emph{NeurIPS 2023}. arXiv:2306.03341.
\bibitem{r35} Sharma, M., Tong, M., Korbak, T., et al. (2023). Towards Understanding Sycophancy in Language Models. arXiv:2310.13548.
\end{thebibliography}
\end{document}